%% file: main.tex
\documentclass[twoside,leqno,twocolumn]{article}

\usepackage[letterpaper]{geometry}

\usepackage{ltexpprt}
\usepackage{hyperref}
\usepackage{adjustbox}
\usepackage{ltexpprt}
\usepackage{hyperref}
\usepackage{amsmath,amsfonts}
\usepackage{mathrsfs}
\usepackage{url}

\usepackage{amssymb}
\usepackage{subfigure}
\usepackage[justification=centering]{caption}
\usepackage{soul}
\usepackage{xcolor}
\usepackage{booktabs}
\usepackage[multiple]{footmisc}
\usepackage{graphicx}
\usepackage{textcomp}
\usepackage[normalem]{ulem}
\def\BibTeX{{\rm B\kern-.05em{\sc i\kern-.025em b}\kern-.08em
    T\kern-.1667em\lower.7ex\hbox{E}\kern-.125emX}}
\makeatletter
\def\@xfootnote[#1]{%
  \protected@xdef\@thefnmark{#1}%
  \@footnotemark\@footnotetext}
\makeatother
\newcommand{\describeContent}[1]{%
\begingroup%
\let\thefootnote\relax%
\footnotetext{#1}%
\endgroup%
}

\begin{document}

\newcommand\relatedversion{}
\renewcommand\relatedversion{\thanks{The full version of the paper can be accessed at \protect\url{https://arxiv.org/abs/1902.09310}}} 

\title{\textbf{Feature Interaction Aware Automated Data Representation Transformation}}
\author{Ehtesamul Azim$^{1}$ \and Dongjie Wang$^{1}$ \and Kunpeng Liu$^2$ \and Wei Zhang$^1$ \and Yanjie Fu$^{3}$ }

\date{}

\maketitle


\fancyfoot[R]{\scriptsize{Copyright \textcopyright\ 2024 by SIAM\\
Unauthorized reproduction of this article is prohibited}}

\let\thefootnote\relax\footnotetext{\scriptsize $^1$ University of Central Florida, Emails:\\ ehtesamul.azim@ucf.edu, dongjie.wang@ucf.edu, wzhang.cs@ucf.edu}
\let\thefootnote\relax\footnotetext{\scriptsize $^2$ Portland State University, Email: kunpeng@pdx.edu}
\let\thefootnote\relax\footnotetext{\scriptsize $^3$ Arizona State University, Email: yanjie.fu@asu.edu}







\input{abstract}
\vspace{-0.6cm}
\input{introduction}

\vspace{-0.2cm}
\input{preliminary}
\vspace{-0.2cm}
\input{method}
\vspace{-0.2cm}
\input{experiments}

\vspace{-0.2cm}
\input{related}

\vspace{-0.2cm}
\input{conclusion}
\section{Acknowledgement}
This research was partially supported by the National Science Foundation (NSF) via the grant numbers: 2040950, 2006889, 2045567, 215230 and 2246796.
\vspace{-0.2cm}
\bibliographystyle{siam}

\bibliography{ref}

\end{document}

%% file: abstract.tex
\begin{abstract}
\small\baselineskip=1pt
Creating an effective representation space is crucial for mitigating the curse of dimensionality, enhancing model generalization, addressing data sparsity, and leveraging classical models more effectively. Recent advancements in automated feature engineering (AutoFE) have made significant progress in addressing various challenges associated with representation learning, issues such as heavy reliance on intensive labor and empirical experiences, lack of explainable explicitness, and inflexible feature space reconstruction embedded into downstream tasks. 
However, these approaches are constrained by: 
1) generation of potentially unintelligible and illogical reconstructed feature spaces, stemming from the neglect of expert-level cognitive processes;
2) lack of systematic exploration, which subsequently results in slower model convergence for identification of optimal feature space.
To address these, we introduce an interaction-aware reinforced generation perspective. We redefine feature space reconstruction as a nested process of creating meaningful features and controlling feature set size through selection.
We develop a hierarchical reinforcement learning structure with cascading Markov Decision Processes to automate feature and operation selection, as well as feature crossing. By incorporating statistical measures, we reward agents based on the interaction strength between selected features, resulting in intelligent and efficient exploration of the feature space that emulates human decision-making. Extensive experiments are conducted to validate our proposed approach. \let\thefootnote\relax\footnotetext{The release code can be found in \url{https://github.com/ehtesam3154/InHRecon}}
\end{abstract}

%% file: introduction.tex
\vspace{-0.1cm}
\section{Introduction}

\begin{figure}[!htbp]
\vspace{-0.2cm}
    \centering
    \includegraphics[width=1.0\linewidth]{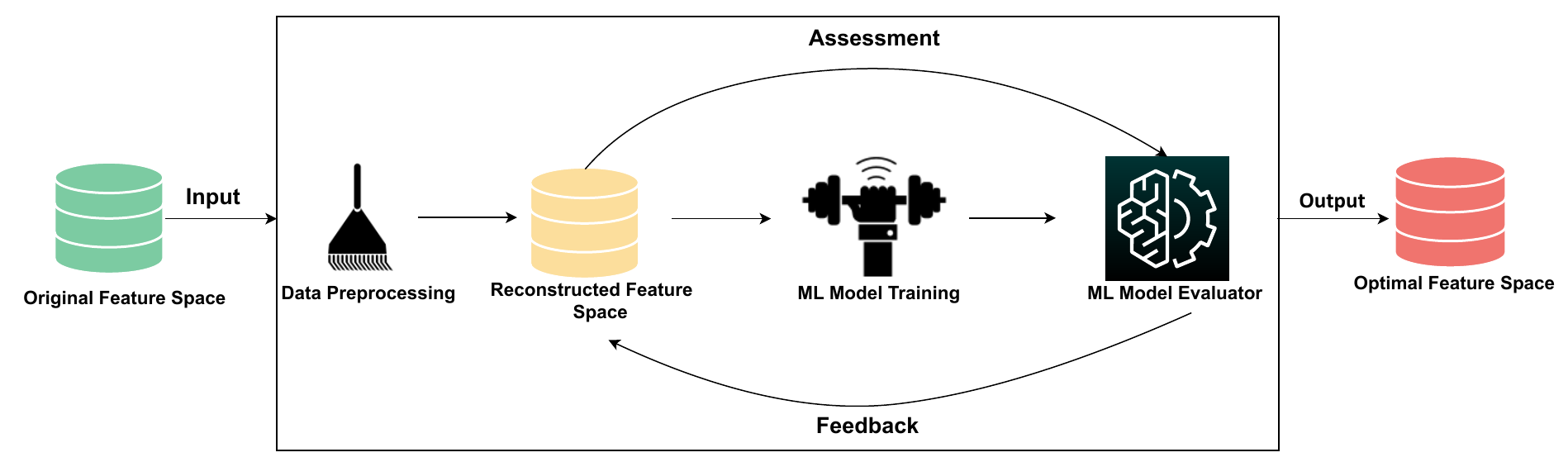}
    \captionsetup{justification=centering}
    \vspace{-0.3cm}
    \caption{We aim to iteratively reconstruct the feature space for an optimal representation space for improved performance in downstream ML task.
    }
    \vspace{-0.5cm}
    \label{fig:autofe_basic}
\end{figure}

With the advent of deep AI, data representation generation has become a key step to the application of machine learning (ML) models. In this work, we investigate the problem of learning to reconstruct an optimal and interpretable feature representation space that enhances performance of a subsequent ML task(e.g, classification, regression) (\textbf{Figure \ref{fig:autofe_basic}}). Formally, given a set of original features, a prediction target, and the specific downstream objective, the goal is to automatically construct an ideal and explainable feature set for said ML task.

Prior literature partially addresses this, starting with feature engineering~\cite{guyon2003introduction,khurana2018feature} to extract a transformed representation of the data. These methods tend to be labor-intensive and have limited ability to automate the extraction.
The next relevant work is representation learning. These include factorization~\cite{fusi2018probabilistic}, embedding~\cite{goyal2018graph} and deep representation learning~\cite{wang2021reinforced}- all of which focus on learning effective latent features. But these often lack interpretability, which limit their deployment in many application scenarios where both high predictive accuracy and a trustworthy understanding of the underlying factors are required.
The final relevant work is learning based feature transformation, which involves traversal transformation graph-based feature generation~\cite{khurana2018feature}, sparsity regularization-based feature selection~\cite{hastie2019statistical} etc. These methods are either deeply integrated into a specific ML model or totally irrelevant to it.
In recent years, significant advancements have been made in automated feature engineering (AutoFE)\cite{chen2019neural, 10.1145/3534678.3539278, horn2019autofeat}. These approaches aim to tackle the challenges associated with reducing the dependence on manual feature engineering. Researchers have also emphasized the importance of ensuring traceable representation space\cite{10.1145/3534678.3539278}, while simultaneously ensuring the flexibility of the reconstructed representation space for any given predictor.

\begin{figure}[!htbp]
\vspace{-0.2cm}
    \centering
    \includegraphics[width=1.0\linewidth]{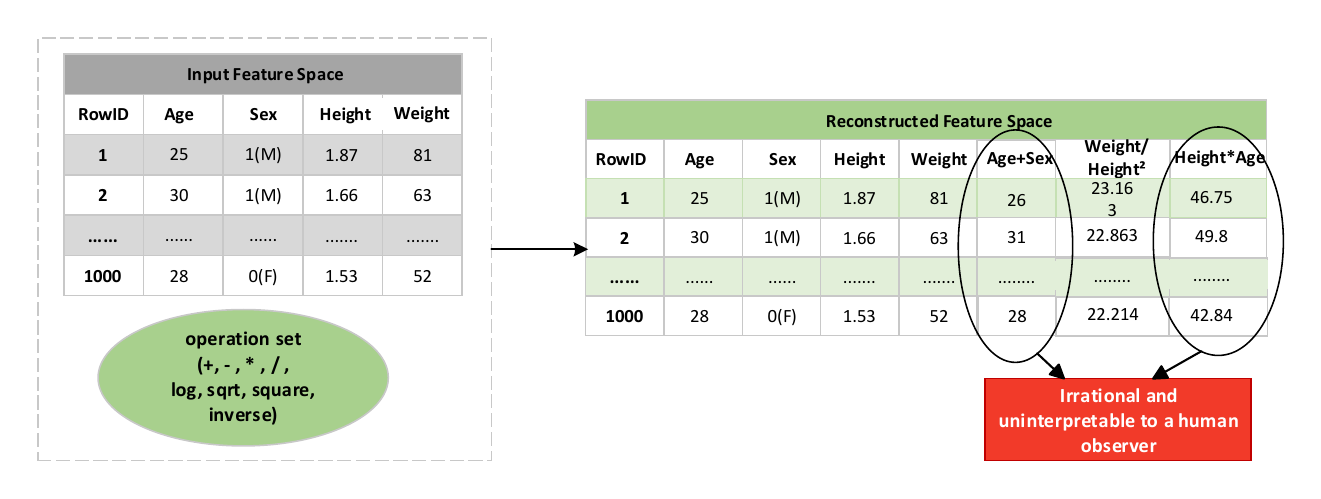}
    \captionsetup{justification=centering}
    \vspace{-0.3cm}
    \caption{One major drawback of existing AutoFE methods: generation of irrational features.
    }
    \vspace{-0.75cm}
    \label{fig:autofe_issue}
\end{figure}

Nevertheless, the mentioned works face two common issues. Classical feature engineering involves expert-driven feature extraction, whereas AutoFE mainly focuses on model-driven feature optimization. While the generated higher-order features might help achieve better performance, many of them are incomprehensible by human observers (\textbf{Figure \ref{fig:autofe_issue}}). \textbf{Issue 1 (expert-level cognition):} \textit{How can we guarantee that the representation space reconstruction yields human-understandable features?} Another issue observed is the statistical insignificance of feature interactions, leading to inefficient exploration of the feature space. This implies that the interactions between some features may have limited impact on the overall predictive performance and the exploration process may focus on feature combinations with minimal contribution to model improvement. \textbf{Issue 2 (systematic exploration):} \textit{how can we ensure a methodical exploration of the feature space during representation space reconstruction for faster convergence?} Our objective is to develop a fresh perspective to address these two well-known yet underexplored challenges to reconstructing an optimal and interpretable representation space. 


\textbf{Our Contributions: An Interaction-aware Reinforced Generation Perspective.} We approach representation space reconstruction through the lens of reinforcement learning (RL). We show that learning to reconstruct is achievable by an interactive process of nested feature generation and selection- where the former focuses on generating new comprehensible features while the later controls feature space size. We emphasize that human intuition and domain expertise in feature engineering can be formulated as machine-learnable policies. We demonstrate that the iterative sequential feature generation can be generalized as a RL task. We find that by expanding the operational capabilities of RL agents and ensuring their statistical awareness, we increase the likelihood of generating interpretable and meaningful variables in the new representation space. Additionally, we demonstrate that we can enhance learning efficiency by rewarding agents based on their level of human-like or statistical awareness.

\textbf{Summary of Proposed Approach.} Based on our findings, we develop a comprehensive and systematic framework to learn a representation space reconstruction policy that can 1) \textbf{Goal 1: explainable explicitness:} provide traceable generation of new features while ensuring their interpretability to human observer, 2) \textbf{Goal 2: self optimization with human-like and statistical awareness:} automatically generate an optimal feature set for a downstream ML task without prior domain knowledge while being aware in both human-like and statistical manner,  \textbf{3) Goal 3: enhanced efficiency and reward augmentation:} enhance the generation and exploration speed in large feature space and augment reward incentive signal to learn clear policy. To accomplish Goal 1, we introduce an iterative strategy involving feature generation and selection which enhances interpretability by allowing us to assign semantic labels to new features and track generation process. To achieve Goals 2 and 3, we break down feature generation process into three distinct Markov Decision Processes(MDPs) to select an operation and two meta feature. To enhance the coordination and learning capabilities of the agents involved, we employ a hierarchical agent architecture that enables state sharing between agents and facilitates development of improved selection policies. To avoid generating uninterpretable features, we design a model structure that can handle both numerical and categorical features. Additionally, we formulate reward functions to incentivize agents based on their selection of operations or feature type, thereby promoting human-level awareness. To ensure statistical awareness, we incorporate H-statistics~\cite{friedman2008predictive}. It measures feature interaction strength by quantifying the extent to which the variation in the prediction of target label depends on selected features' interaction. This approach enhances efficiency in large feature spaces, providing clearer guidance for selection policies.

%% file: preliminary.tex
\section{Defintions and Problem Formulation}
\begin{figure*}[!h]
\vspace{-0.2cm}
    \centering
    \includegraphics[width=1.0\linewidth]{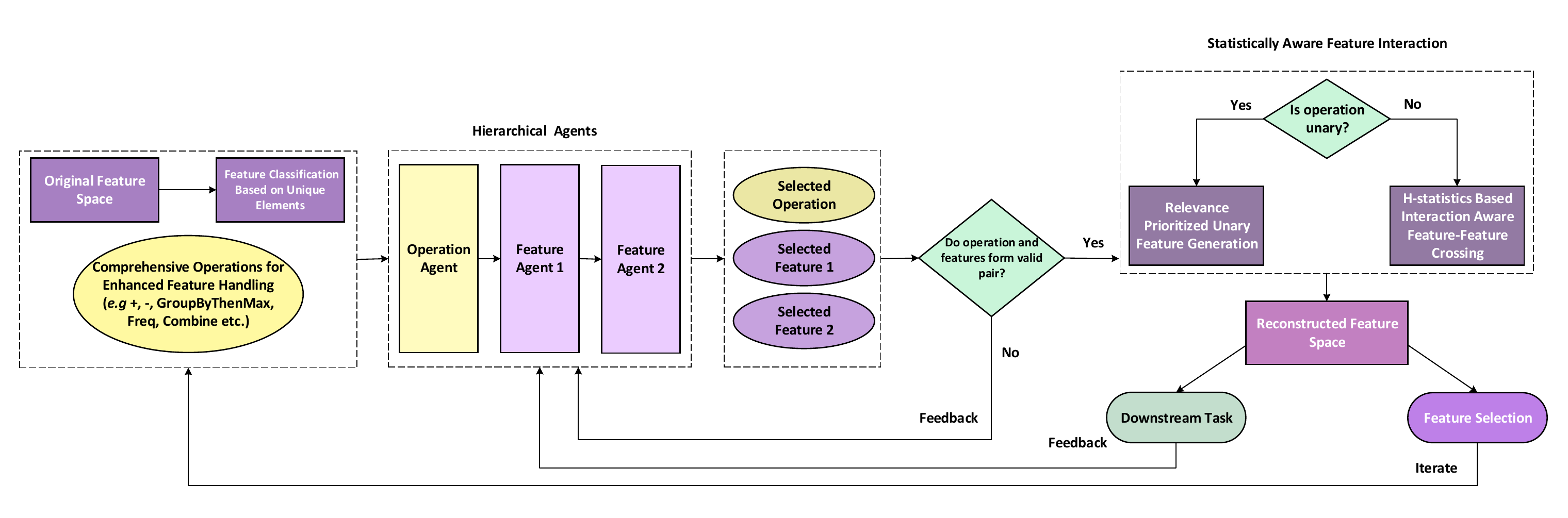}
    \vspace{-0.2cm}
    \captionsetup{justification=centering}
    \vspace{-0.33cm}
    \caption{Overview of the proposed framework. Feature classification step categorizes features into continuous and categorical types, along with an enhanced operation set. Hierarchical agents select an operation and two features, followed by statistically aware feature interaction to generate new features. Responsible agents are penalized for invalid operation-feature pairs. The updated feature set evaluated in a downstream task. Feature selection is applied to control the feature set size, with iterations continuing until optimization or set limit.
    }
    \vspace{-0.5cm}
    \label{fig:framework}
\end{figure*}

\noindent\textbf{Operation Set.}
We apply mathematical operations to existing features to generate new ones. Previous AutoFE studies typically treat all feature columns as numerical and restrict the operation set to numerical operators. To effectively handle both numerical and categorical features, we extend our operation set $\mathcal{O}$ to include 26 operators, providing a wider range of functionalities, such as ``Combine", ``GroupByThenMin", ``Freq"  etc.
\vspace{0.1cm}

\noindent\textbf{Hierarchical Agent.}
To address automated feature generation, we introduce a hierarchical agent structure with three agents: one operation agent and two feature agents. These agents collaborate in dividing the feature generation problem into two sub-problems: operation selection and candidate feature selection, working together by sharing state information. 

\noindent\textbf{Problem Statement.}
Our research aims to learn an optimal and meaningful representation space for improved performance in a downstream ML task. Formally, given a dataset $D<\mathcal{F},y>$ with a feature set $\mathcal{F}$ and a target label $y$, an operator set $\mathcal{O}$, and a downstream ML task $A$ (e.g., classification, regression), our goal is to automatically reconstruct an optimal and interpretable feature set $\mathcal{F}^{*}$ that maximizes:
\begin{equation}
\label{objective}
    \mathcal{F}^{*} = argmax_{\mathcal{\hat{F}}}(V_A(\mathcal{\hat{F}}, y))
\end{equation}
$\mathcal{\hat{F}}$ denotes subset comprising combinations of the original feature set $\mathcal{F}$ and the generated features $\mathcal{F}_{g}$, which are obtained by applying operation set $\mathcal{O}$ to original feature set using a specific algorithmic structure.


%% file: method.tex
\section{Methodology}
An overview of our proposed framework, \textbf{In}teraction-aware \textbf{H}ierarchical \textbf{R}einforced Feature Space \textbf{Recon}struction(\textbf{InHRecon}) is illustrated in \textbf{Figure \ref{fig:framework}}. We initiate the process by classifying the feature space into two categories: categorical and numerical. This classification is based on the number of unique elements in each feature column. Our approach employs an operation-feature-feature strategy to combine two existing features at each step. The technical details of each of rest of the component are discussed below.

\vspace{-0.4cm}
\subsection{Hierarchical Reinforced Feature Selection and Generation}
We devise a hierarchical RL agent structure framework for automated feature generation based on two key findings. \textbf{Firstly}, we emphasize that programming optimal selection criteria for features and operations can be treated as machine-learnable policies, addressed by three learning agents. \textbf{Secondly}, we observe that the agents operate in a hierarchical manner, with interconnected sequential decision-making process. Within each iteration, the agents divide the feature generation problem into sub-problems of selecting operation and selecting feature(s). They make decisions sequentially, where the choices made by an upstream agent have an impact on the state of the environment for downstream agents. 

\noindent\textbf{Three utility metrics for reward quantification.} We propose three metrics to quantify feature usefulness, and form three MDPs to learn selection policies.

\noindent\textit{\uline{Metric 1: Validity of operation.}} We ensure operation validity by considering the compatibility between the selected feature types and operations. For instance, applying a mathematical operator like `sqrt' to a categorical feature such as `Sex' is not appropriate. To address such instances, we hold the responsible agent accountable through this metric and encourage improved future feature choices. We use the notation ${U}(f_t|o_t)$ to represent the metric associated with a selected operation-feature pair $(o_t, f_t)$ at $t$-th iteration. 


\noindent\textit{\uline{Metric 2: Interaction strength of selected features.}}
To enhance statistical awareness and efficient exploration by agents, we consider how feature interaction influences the outcome of the target label. This interaction strength is quantified using H-statistics, the details of which is given in section \ref{h_stats}. 

\noindent\textit{\uline{Metric 3: Downstream Task Performance.}} We evaluate the effectiveness of the feature set in downstream task, such as regression or classification, using a utility metric such as 1-RAE, Precision, Recall, or F1 score.

\noindent\textbf{Learning Selection Policy for Operation and Features.} Leveraging these metrics, we develop three MDPs to learn three agent polices to select the best operation and feature(s). We adopt the $t$-th iteration as an example to illustrate the calculation process.

\begin{figure}[!htbp]
\vspace{-0.2cm}
    \centering
    \includegraphics[width=1.0\linewidth]{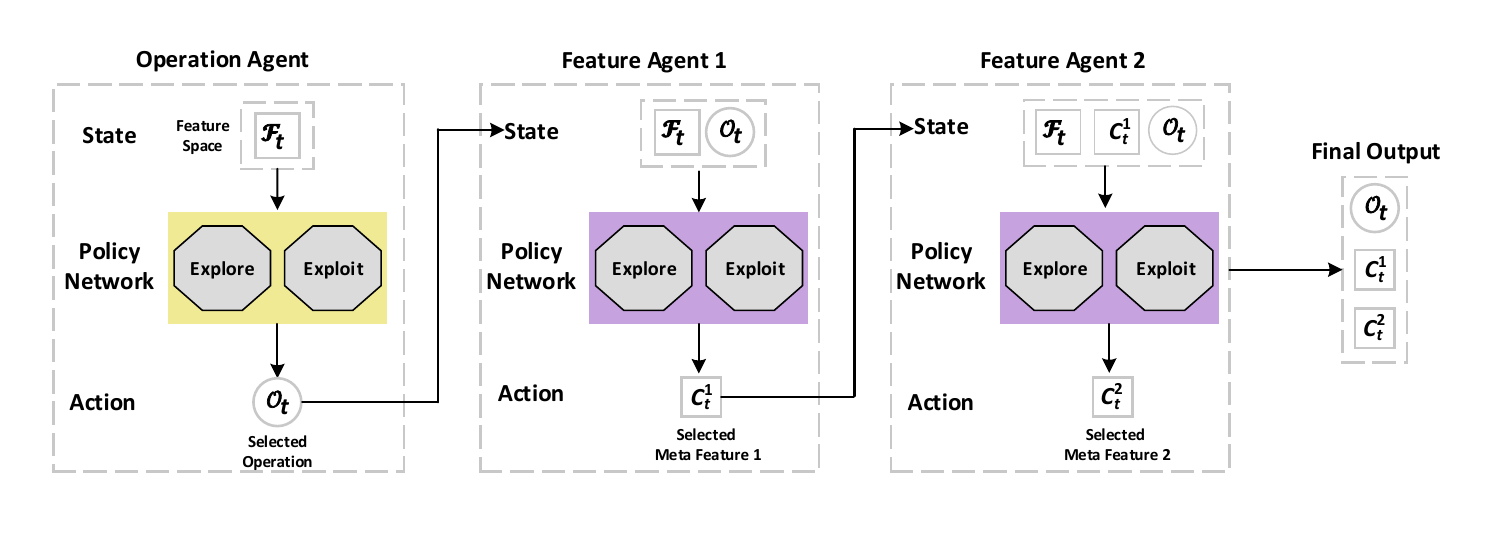}
    \vspace{-0.2cm}
    \captionsetup{justification=centering}
    \vspace{-0.5cm}
    \caption{Proposed hierarchical agent structure}
    \vspace{-0.6cm}
    \label{fig:hrl}
\end{figure}
\smallskip

\noindent\uline{Learning Selection Policy for Operation Agent.} The operation agent picks the best operation from an operation set as a feature crossing tool. Its learning system includes: \textbf{i) State:} its state is an embedding of the generated feature set of the previous iteration. Let $Rep$ be a state representation method(discussed later), the state can be denoted by $s_t^1 = Rep(\mathcal{F}_{t-1})$. $\mathcal{F}_{t-1}$ is the current feature space observed by the agent. \textbf{ii) Action:} its action is the selected operation, denoted by $a_t^o = o_t$.  \textbf{iii) Reward:} its reward is performance improvement on downstream task, denoted by $\mathcal{R}(s_t^o,a_t^o) = V_{A_t} - V_{A_{opt}}$, where $V_{A_t}$ is model performance after the $t$-th iteration with $V_{A_{opt}}$ being the best performance achieved so far.

\noindent\uline{Learning Selection Policy for Feature Agent 1.} This agent selects the first meta feature. Its learning system includes: \textbf{i) State:} its state is the combination of $Rep(\mathcal{F}_{t-1})$ and vectorized representation of the operation selected by operation agent, denoted by $s^1_t = Rep(\mathcal{F}_{t-1}) \oplus Rep(o_t)$, where $\oplus$ indicates concatenation. \textbf{ii) Action:} its action is the first meta feature selected from the observed feature space, denoted by $a_t^1 = f^1_t$. \textbf{iii) Reward:} its reward is determined by the utility score of the selected feature and the improvement in the downstream task performance. The reward can be formulated as $\mathcal{R}(s_t^1,a_t^1) = {U}(f^1_t|o_t) + (V_{A_t} - V_{A_{opt}})$.

\noindent\uline{Learning Selection Policy for Feature Agent 2.} This agent selects the best meta feature 2. Its learning system includes: \textbf{i) State:} The combination of $Rep(\mathcal{F}_{t-1})$, $Rep(f_{t}^1)$ and vectorized representation of the operation, denoted by $s_t^2 = Rep(\mathcal{F}_{t-1})\oplus Rep(f_{t-1}^1) \oplus Rep(o_t)$  \textbf{ii) Action:} The meta feature 2 selected from the observed feature space, denoted by $a_t^2 = f^2_t$. \textbf{iii) Reward:} includes the utility score of the selected feature, the interaction strength between features $f^1_t$ and $f^2_t$, $H_{f^1_t, f^2_t}$ measured by H-statistics and improvement in downstream task performance. 
We formulate the reward as $\mathcal{R}(s_t^2,a_t^2) = {U}(f^2_t|o_t) + H_{f^1_t, f^2_t} + (V_{A_t} - V_{A_{opt}})$.

\smallskip
\noindent\textbf{State Representation of Feature Space and Operation.} We propose to map the observed feature space to a vector to characterize its state. Given feature space $\mathcal{F}$, we compute its descriptive statistics(\textit{i.e} count, standard deviation, min, max and first, second and third quartile) for each of its column. We then calculate the descriptive statistics of the outcome of this step for each row and thus, obtain descriptive matrix of shape $\mathbb{R}^{7\times 7}$. The descriptive matrix is flattened to obtain the final representation $Rep(\mathcal{F})\in \mathbb{R}^{1\times 49}$ of the feature space. With this representation method, we generate a fixed-size state vector that adapts to the varying size of the feature set at each iteration. For the operators, we employ one-hot encoding, represented by $Rep(o)$. 
 \begin{figure}[!htbp]
 \vspace{-0.2cm}
    \centering
    \includegraphics[width=1.0\linewidth]{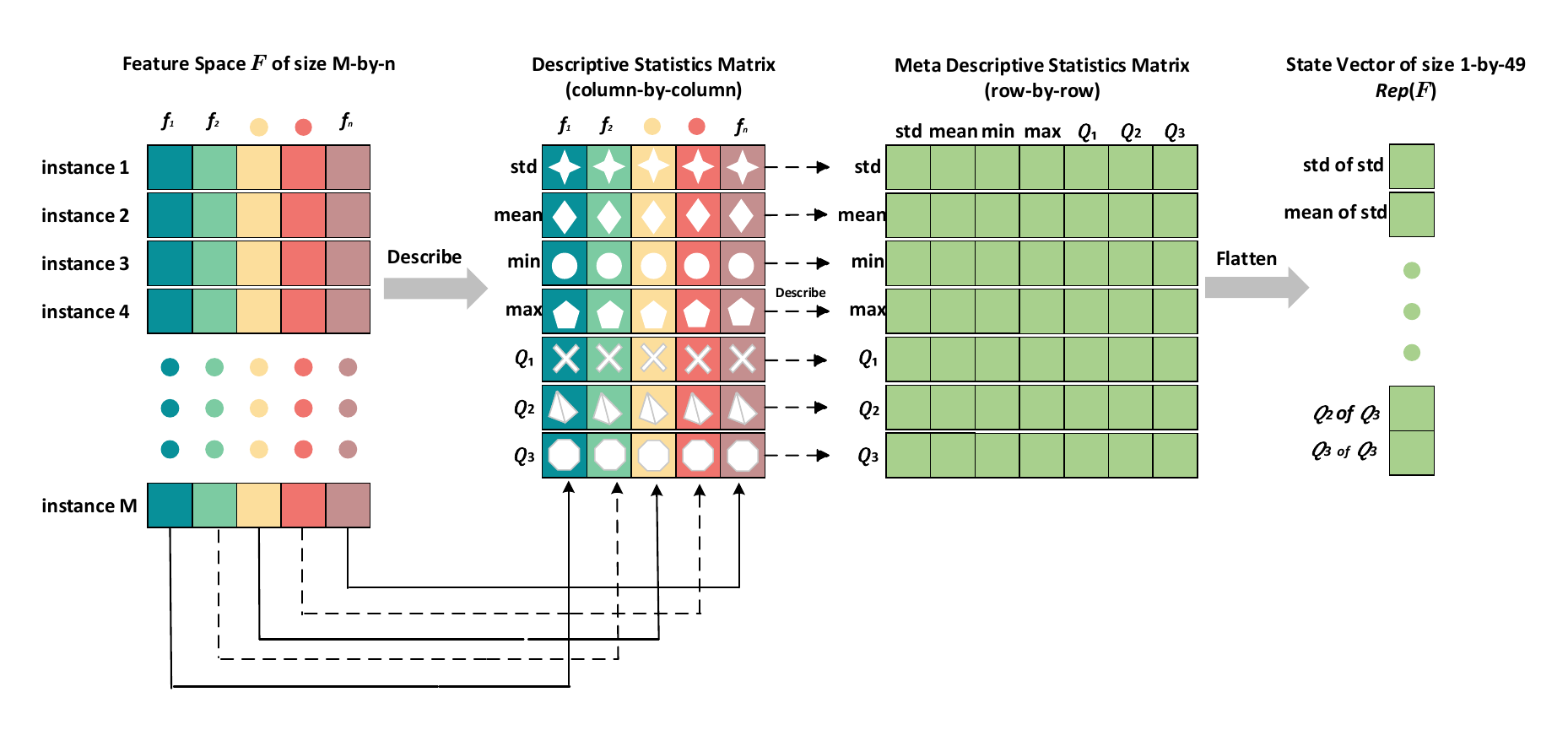}
    \vspace{-0.3cm}
    \captionsetup{justification=centering}
    \vspace{-0.2cm}
    \caption{Illustration of state representation extraction}
    \vspace{-0.5cm}
    \label{fig:state_repr}
\end{figure}

\smallskip
\noindent\textbf{Solving the Optimization Problem.} During the iterative feature generation process, we train our agents to maximize the discounted and cumulative reward. To achieve this, we minimize the temporal difference error $\mathcal{L}$ converted from the Bellman equation, given by:
\begin{equation}
\label{q_update}
    \mathcal{L} = Q(s_t, a_t) - (\mathcal{R}(s_t, a_t) + \gamma*
    \text{max}_{a_{t+1}}Q(s_{t+1}, a_{t+1}))
\vspace{-0.2cm}
\end{equation}

where $Q$ is the estimated $Q$-function and $\gamma\in[0\sim1]$ is the discounted factor. After convergence, agents discover the optimal policy $\pi^*$ to choose the most appropriate action (\textit{i.e} feature or operation) based on the state of the Q-value, formulated as follows:
\begin{equation}
\label{rl_policy}
    \pi^*(a_t|s_t) = \text{argmax}_a Q(s_t,a)
\vspace{-0.2cm}
\end{equation}

\vspace{-0.4cm}
\subsection{Feature Generation and Post-processing}
\label{h_stats}
We found that giving our agents human-like statistical awareness can generate more meaningful, interpretable features at an accelerated exploration speed. Based on the selection results, our RL system is faced with two generation scenarios: (1) an operation and two features are selected (2) an operation and a feature are selected. In cases where the selected feature(s) are deemed invalid for the selected operation (e.g.,selected features are `Weight' and `Height' and selected operation is `+' or `GroupByThenMin'), feature generation process for that iteration is bypassed, and responsible agents are penalized based on previously discussed utility metrics.

\textbf{Scenario 1: H-statistics Based Interaction Aware Feature-Feature Crossing.}
Existing AutoFE literature solely focuses on generating higher-order features to improve downstream task performance. However, this trial-and-error approach lacks efficiency and does not align with human expert intuition. To make our model efficient by giving it statistical awareness, we incorporate Friedman’s H-statistics~\cite{friedman2008predictive}. We here present two-way interaction measure that tells us whether and to what extent two features in the model interact with each other. When two features don’t interact, we can decompose partial dependence(PD) function as follows(assuming PD functions are centered at zero):
\begin{equation}
\label{pd_1}
    PD_{jk}(f_j,f_k) = PD_{j}(f_j) + PD_{k}(f_k)
\vspace{-0.1cm}
\end{equation}
where $PD_{j}(f_j)$ and $PD_{k}(f_k)$ the PD functions of respective features and $PD_{jk}(f_j,f_k)$ is their 2-way PD function. PD functions measure the marginal effect a feature has on the predicted outcome of a ML model. For regression, the PD function can be defined as,
\begin{equation}
\label{pd_2}
    PD_{s}(x_s) = E[\hat{f}(x_s,X_C)] = \int\hat{f}(x_s,X_C)d\mathbb{P}(X_C)
\vspace{-0.3cm}
\end{equation}
${x_s}$ being the features for which partial dependence is to be calculated and ${X_c}$ the other features used in ML model $\hat{f}$. Expected value $E$ is over the marginal distribution of of all variables $X_c$ not represented in $x_s$. PD works by marginalizing the ML model output over the distribution of the features in set $C$, so that the function shows the relationship between the features in set $S$ and the predicted outcome.

Equation \ref{pd_1} expresses the PD function without interactions between ${f_j}$ and ${f_k}$. The observed PD function is compared to the no-interaction decomposition, and the difference represents the interaction strength. The variance of the PD output quantifies the interaction between the two features. An interaction statistic of 0 indicates no interaction, while a statistic of 1 suggests that the prediction relies solely on the interaction. Mathematically, the H-statistic proposed by Friedman and Popescu for the interaction between ${f_j}$ and ${f_k}$ is:
\begin{equation}
    \label{h_stat}
    H_{j,k} = \frac{\sum_{i=1}^{n}[PD_{jk}(f_{j}^{(i)}, f_{k}^{(i)}) - PD_{j}(f_{j}^{(i)}) - PD_{k}(f_{k}^{(i)})]^2}{\sum_{i=1}^{n} PD^{2}_{jk}(f_{j}^{(i)}, f_{k}^{(i)})}
\vspace{-0.2cm}
\end{equation}
where n is the number of instances in the dataset. For ease of understanding, we have refrained from delving into the detailed calculation for interactions involving more than two features, which we implement in our work. In our implementation, rather than directly computing the interaction strength between two higher-order features, we calculate it for their respective `parent' features.  Our rationale behind this approach is that although the interaction strength between two features may be relatively low, exploring the interactions between their parent features can yield valuable insights and justify the agents' exploration efforts.

\textbf{Scenario 2: Relevance Prioritized Unary Feature Generation.} Inspired by~\cite{10.1145/3534678.3539278}, we directly apply the operation to the feature that is more relevant to target label when an unary operation and two features are selected. 
We measure relevance using mutual information(MI) between feature $f\in\mathcal{F}$ and target label $y$, quantified by:
$
rel = MI(f,y)
$. The operation is applied to the more relevant feature to generate new feature.

\textbf{Post-generation Processing.}
After generating new features, we combine them with the original features to create an updated feature set and evaluate the predictive performance on downstream task. The performance serves as feedback to update the agents' policies for the next round of feature generation. To control feature set size, we apply feature selection if the size surpasses a threshold, using K-best feature selection. The tailored feature set becomes the original feature set for the next iteration. 
Upon reaching the maximum number of iteration, the algorithm concludes by returning the optimal feature set $\mathcal{F^{*}}$.

\begin{table*}[!h]
\centering
\caption{Overall performance comparison. `C' for classification and `R' for regression.}
\vspace{-3mm}
\label{performance}
\setlength{\tabcolsep}{2.0mm}{\resizebox{\linewidth}{!}{
\begin{tabular}{|c|c|c|c|c|c|c|c|c|c|c|}
\hline
Dataset & Source & C/R & Instances\textbackslash{}Features & ERG & LDA & AFT & NFS & TTG & GRFG & InHRecon \\ \hline
Higgs Boson & UCIrvine & C & 50000\textbackslash{}28 & 0.674 & 0.509 & 0.711 & 0.715 & 0.705 & 0.716 & \textbf{0.718} \\ \hline
Amazon Employee & Kaggle & C & 32769\textbackslash{}9 & 0.740 & 0.920 & 0.943 & 0.935 & 0.806 & 0.946 & \textbf{0.947} \\ \hline
PimaIndian & UCIrvine & C & 768\textbackslash{}8 & 0.703 & 0.676 & 0.736 & 0.762 & 0.747 & 0.767 & \textbf{0.778} \\ \hline
SpectF & UCIrvine & C & 267\textbackslash{}44 & 0.748 & 0.774 & 0.775 & 0.842 & 0.788 & 0.854 & \textbf{0.878} \\ \hline
SVMGuide3 & LibSVM & C & 1243\textbackslash{}21 & 0.747 & 0.683 & 0.829 & 0.831 & 0.766 & 8.842 & \textbf{0.850} \\ \hline
German Credit & UCIrvine & C & 1001\textbackslash{}24 & 0.661 & 0.627 & 0.751 & 0.765 & 0.731 & 0.769 & \textbf{0.773} \\ \hline
Credit Default & UCIrvine & C & 30000\textbackslash{}25 & 0.752 & 0.744 & 0.799 & 0.799 & 0.809 & 0.800 & \textbf{0.812} \\ \hline
Messidor Features & UCIrvine & C & 1150\textbackslash{}19 & 0.635 & 0.580 & 0.679 & 0.746 & 0.726 & \textbf{0.757} & 0.738 \\ \hline
Wine Quality Red & UCIrvine & C & 999\textbackslash{}12 & 0.611 & 0.600 & 0.658 & 0.666 & 0.647 & 0.686 & \textbf{0.706} \\ \hline
Wine Quality White & UCIrvine & C & 4900\textbackslash{}12 & 0.587 & 0.571 & 0.673 & 0.679 & 0.638 & 0.685 & \textbf{0.696} \\ \hline
SpamBase & UCIrvine & C & 4601\textbackslash{}57 & 0.931 & 0.908 & 0.951 & 0.955 & 0.959 & 0.958 & \textbf{0.971} \\ \hline
AP-omentum-ovary & OpenML & C & 275\textbackslash{}10936 & 0.705 & 0.117 & 0.783 & 0.804 & 0.795 & 0.808 & \textbf{0.811} \\ \hline
Lymphography & UCIrvine & C & 148\textbackslash{}18 & 0.638 & 0.737 & 0.833 & 0.859 & 0.846 & 0.866 & \textbf{0.875} \\ \hline
Ionosphere & UCIrvine & C & 351\textbackslash{}34 & 0.926 & 0.730 & 0.827 & 0.942 & 0.938 & 0.946 & \textbf{0.954} \\ \hline
Bikeshare DC & Kaggle & R & 10886\textbackslash{}11 & 0.980 & 0.794 & 0.992 & 0.991 & 0.991 & 0.992 & \textbf{0.994} \\ \hline
Housing Boston & UCIrvine & R & 506\textbackslash{}13 & 0.617 & 0.174 & 0.641 & 0.654 & 0.658 & 0.658 & \textbf{0.660} \\ \hline
Airfoil & UCIrvine & R & 1503\textbackslash{}5 & 0.732 & 0.463 & 0.774 & 0.771 & 0.783 & 0.787 & \textbf{0.793} \\ \hline
Openml\_618 & OpenML & R & 1000\textbackslash{}50 & 0.427 & 0.372 & 0.665 & 0.640 & 0.587 & 0.668 & \textbf{0.673} \\ \hline
Openml\_589 & OpenML & R & 1000\textbackslash{}25 & 0.560 & 0.331 & 0.672 & 0.711 & 0.682 & \textbf{0.739} & 0.723 \\ \hline
Openml\_616 & OpenML & R & 500\textbackslash{}50 & 0.372 & 0.385 & 0.585 & 0.593 & 0.559 & 0.603 & \textbf{0.605} \\ \hline
Openml\_607 & OpenML & R & 1000\textbackslash{}50 & 0.406 & 0.376 & 0.658 & 0.675 & 0.639 & \textbf{0.680} & 0.671 \\ \hline
Openml\_620 & OpenML & R & 1000\textbackslash{}25 & 0.584 & 0.425 & 0.663 & 0.698 & 0.656 & \textbf{0.714} & 0.694 \\ \hline
Openml\_637 & OpenML & R & 500\textbackslash{}50 & 0.497 & 0.494 & 0.564 & 0.581 & 0.575 & 0.589 & \textbf{0.625} \\ \hline
Openml\_586 & OpenML & R & 1000\textbackslash{}25 & 0.546 & 0.472 & 0.687 & 0.748 & 0.704 & \textbf{0.763} & 0.756 \\ \hline
\end{tabular}%
}}
\vspace{-3mm}
\end{table*}

\begin{figure*}[!h]
\centering
\subfigure[PimaIndian]{
\includegraphics[height=3.5cm,width=4cm]{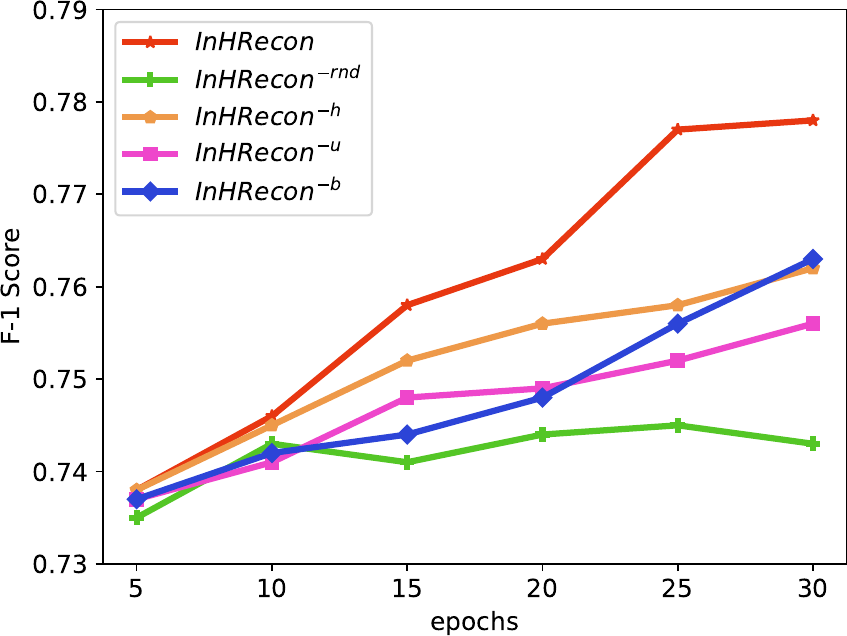}
}
\hspace{-3mm}
\subfigure[Credit Default]{ 
\includegraphics[height=3.5cm,width=4cm]{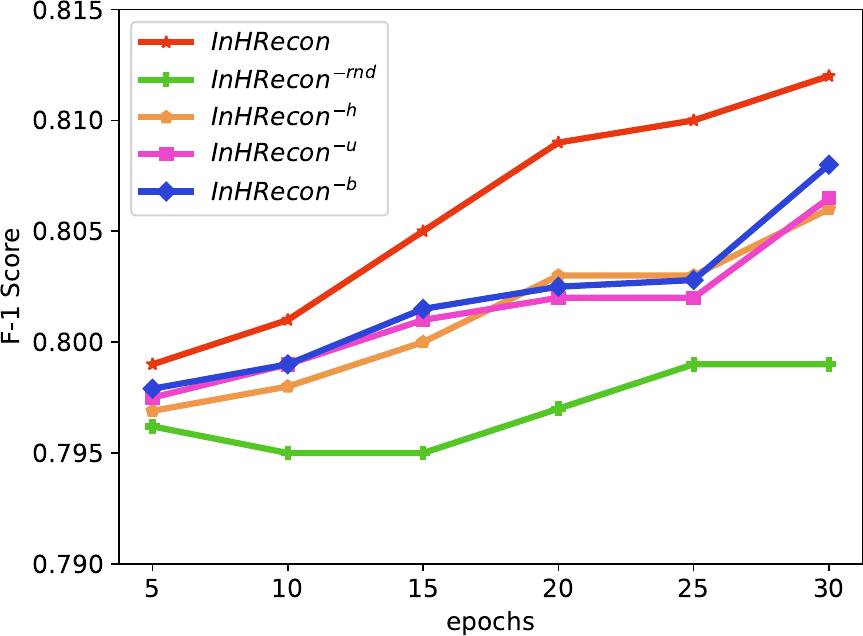}
}
\hspace{-3mm}
\subfigure[Housing Boston]{
\includegraphics[height=3.5cm,width=4cm]{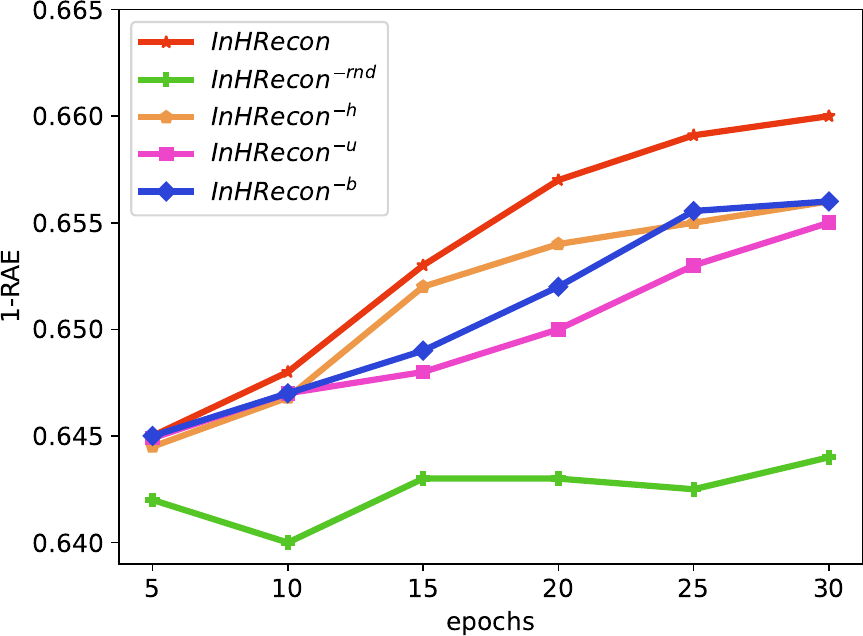}
}
\hspace{-3mm}
\subfigure[Openml\_637]{ 
\includegraphics[height=3.5cm,width=4cm]{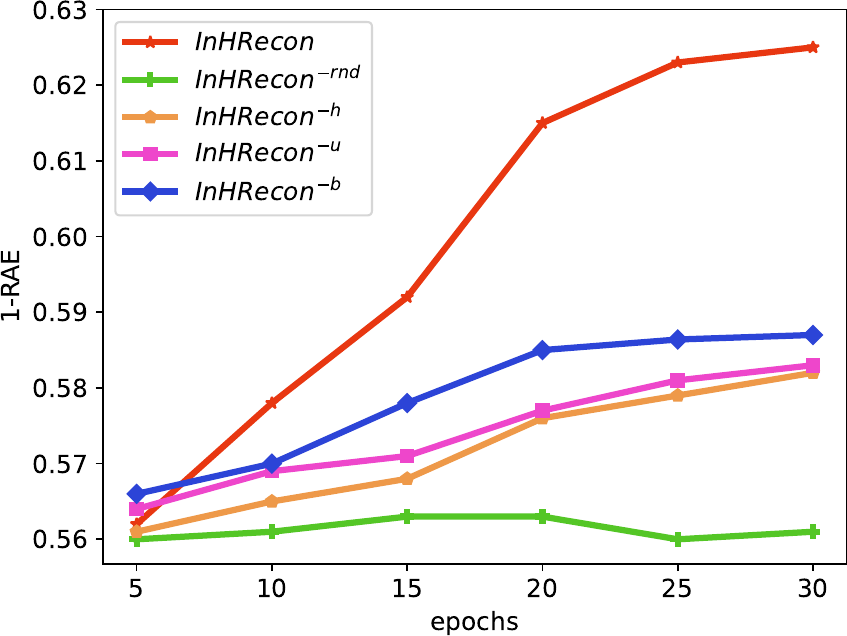}
}
\vspace{-0.35cm}
\caption{Comparison of convergence of different variants of InHRecon}
\label{ab_study}
\vspace{-0.7cm}
\end{figure*}

%% file: experiments.tex
\section{Experiments}

\subsection{Experimental Setup}
24 public datasets from LibSVM\footnote[*]{\url{https://www.csie.ntu.edu.tw/~cjlin/libsvmtools/datasets/}}, UCI\footnote[$\dagger$]{\url{https://archive.ics.uci.edu/}}, Kaggle\footnote[$\ddagger$]{\url{https://www.kaggle.com/datasets}} and OpenML\footnote[$\mathsection$]{\url{https://www.openml.org/}} are utilized  to evaluate InHRecon, including 14 classification and 10 regression tasks. \textbf{Table\ref{performance}} shows the statistics of the data. To evaluate the classification tasks, we use F-1 score. For regression tasks, we use 1-relative absolute error (RAE) to evaluate the accuracy. 

\vspace{-0.4cm}
\subsubsection{Baseline Algorithms} \label{baseline}We compare \textbf{InHRecon} with six state-of-the-art feature generation methods: 
(1) \textbf{ERG} expands the feature space by applying operations to each feature and then does feature selection;
(2) \textbf{LDA}~\cite{blei2003latent} extracts latent features via matrix factorization;
(3)\textbf{AFT}~\cite{horn2019autofeat} is an enhanced ERG implementaion that explores feature space and adopts multi-step feature selection leveraging L1-regularization;
(4) \textbf{NFS}~\cite{chen2019neural} mimics feature transformation trajectory for each feature and optimizes the feature generation process through RL;
(5) \textbf{TTG}~\cite{khurana2018feature} records the feature generation process using a transformation graph and employs RL to explore the graph for the best feature set;
(6) \textbf{GRFG}~\cite{10.1145/3534678.3539278} adopts a groupwise feature generation approach, leveraging MI theory. To validate the impact of each technical component, we developed four variants of  InHRecon: 
(i) \textbf{InHRecon$^{-rnd}$} randomly picks an operation and feature(s);
(ii) \textbf{InHRecon$^{-h}$} treats all feature columns as numerical;
(iii) \textbf{InHRecon$^{-u}$} randomly selects a feature from the selected features when operator is unary;
(iv) \textbf{InHRecon$^{-b}$} doesn't utilize H-statistics for binary operations.
We adopted random forest as the downstream ML model and a 5-fold stratified cross-validation approach. 
\begin{figure*}[htbp]
\centering
\subfigure[PimaIndian]{
\includegraphics[height=3.5cm, width=4cm]{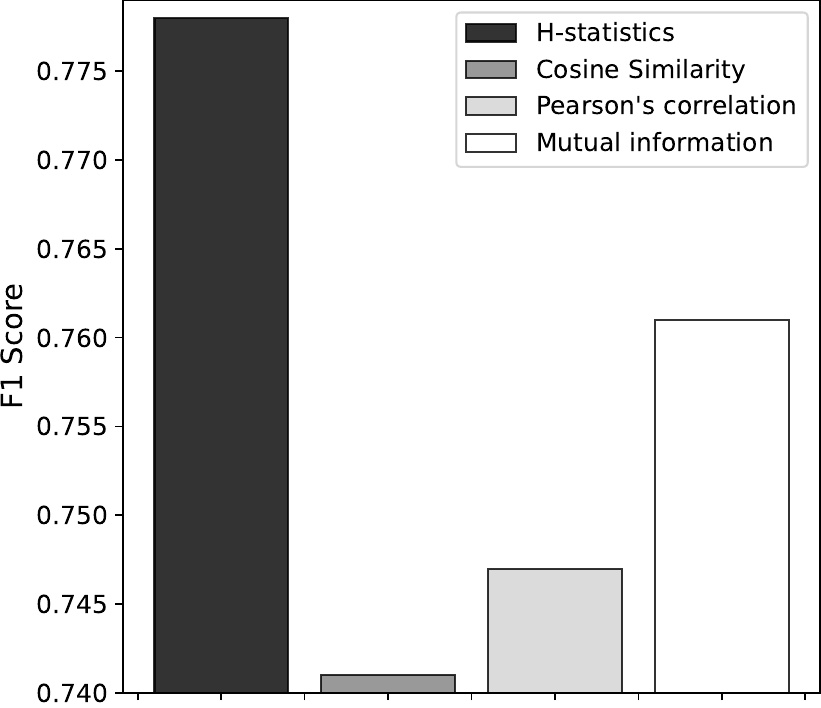}
}
\hspace{-3.5mm}
\subfigure[Credit Default]{ 
\includegraphics[height=3.5cm, width = 4cm]{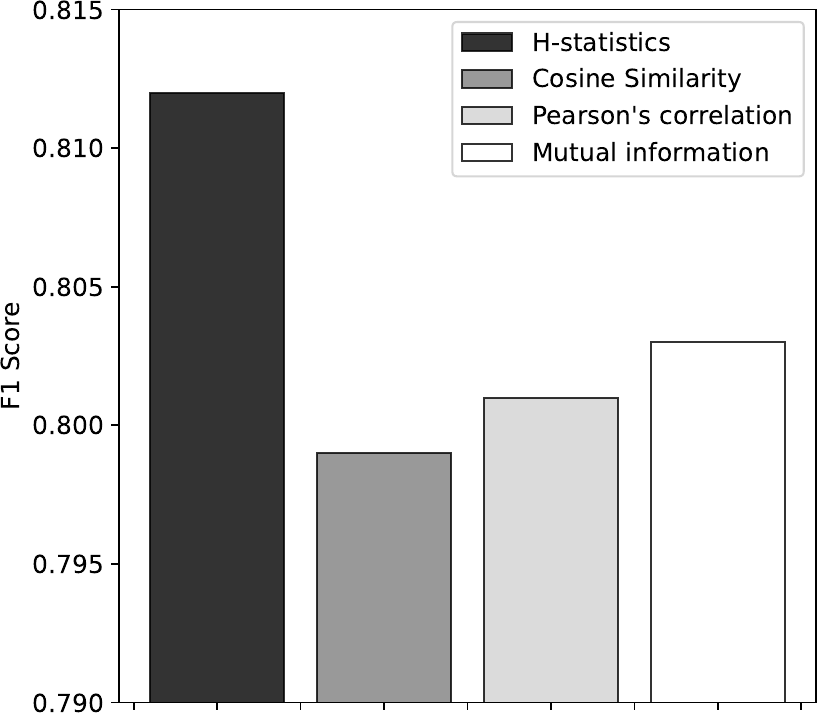}
}
\hspace{-3.5mm}
\subfigure[Housing Boston]{
\includegraphics[height=3.5cm, width=4cm]{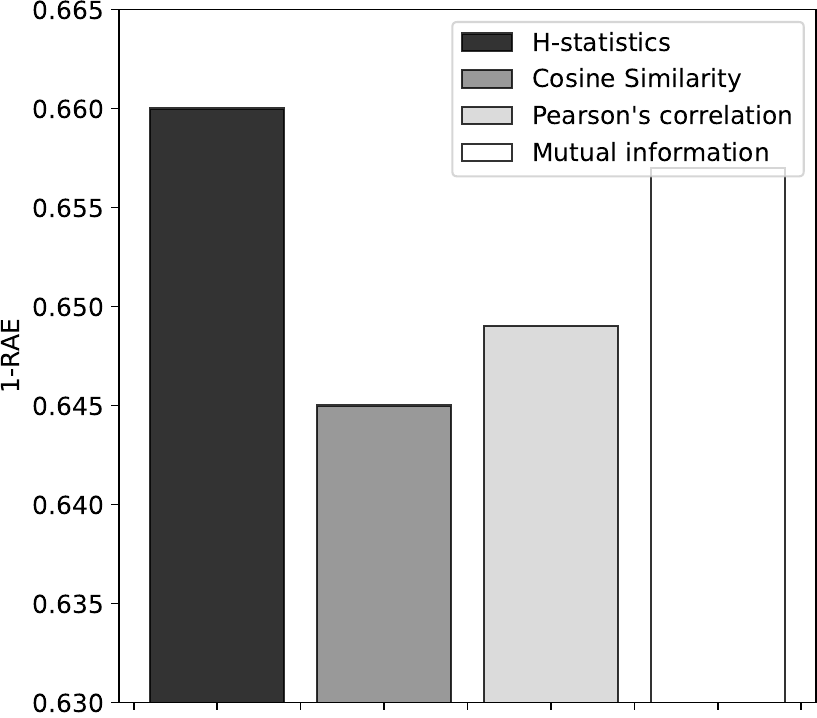}
}
\hspace{-3.5mm}
\subfigure[Openml\_637]{ 
\includegraphics[height=3.5cm, width=4cm]{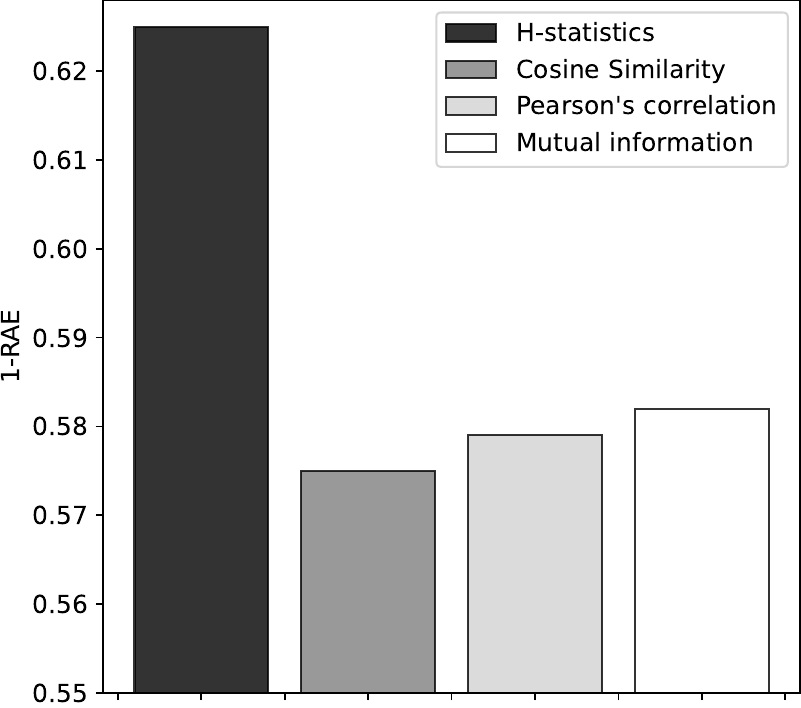}
}
\vspace{-0.2cm}
\caption{Comparison of different interaction strategy in terms of F1 or 1-RAE.}
\label{interact}
\vspace{-0.35cm}
\end{figure*}

\begin{figure*}[htbp]
\centering
\subfigure[PimaIndian]{
\includegraphics[height=3.5cm, width=4cm]{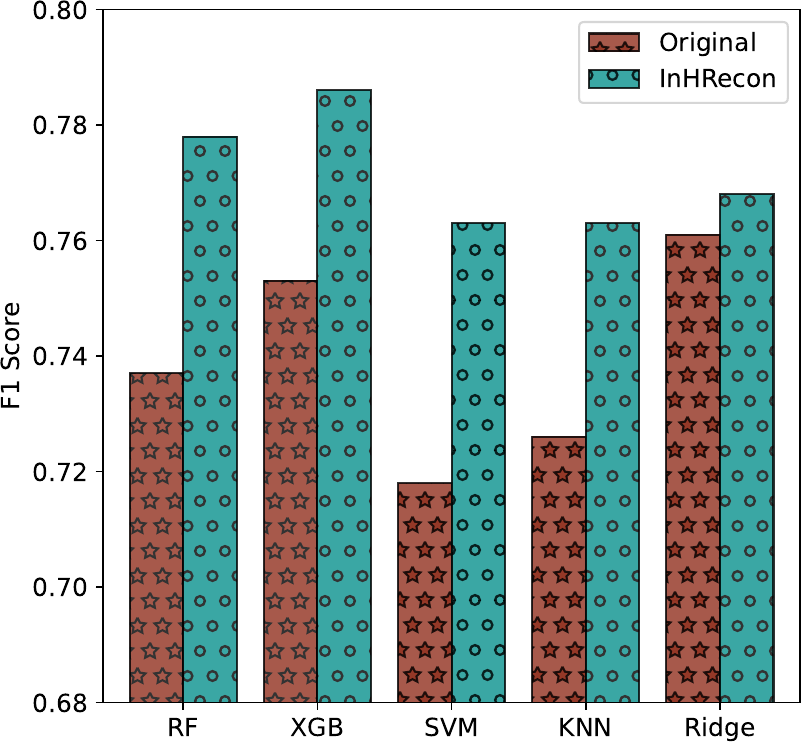}
}
\hspace{-3.5mm}
\subfigure[Credit Default]{ 
\includegraphics[height=3.5cm, width = 4cm]{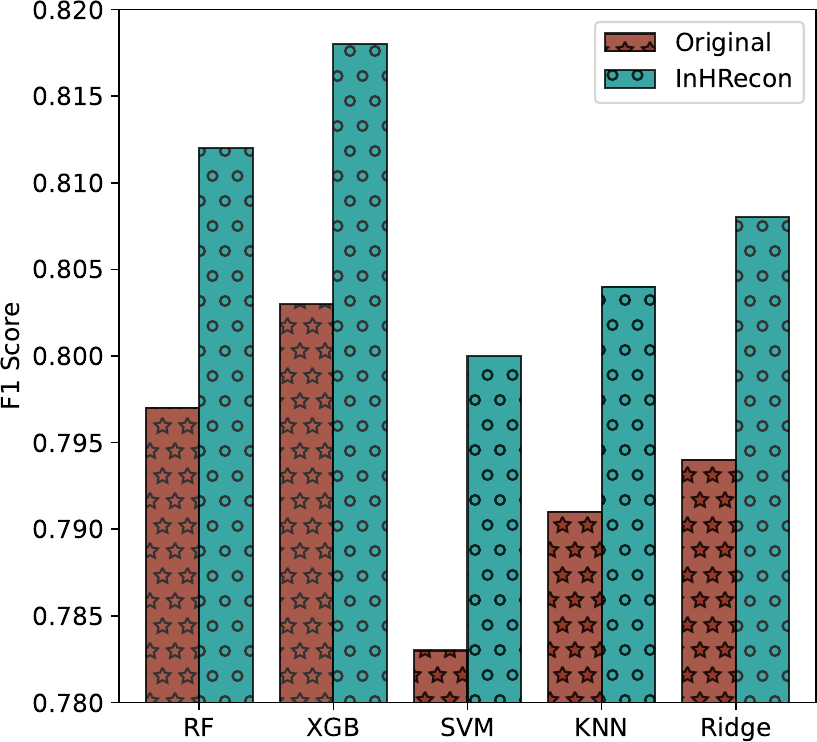}
}
\hspace{-3.5mm}
\subfigure[Housing Boston]{
\includegraphics[height=3.5cm, width=4cm]{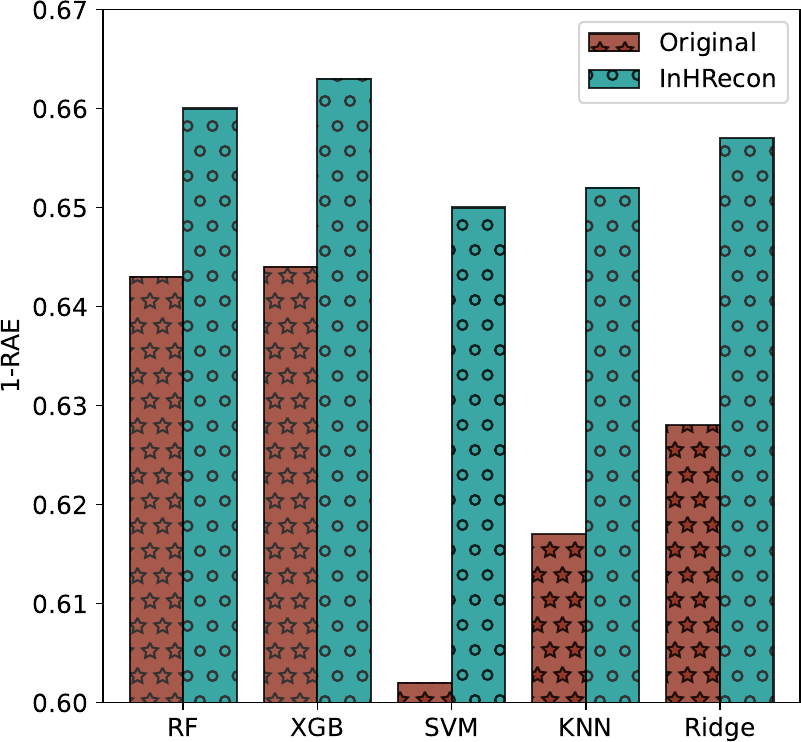}
}
\hspace{-3.5mm}
\subfigure[Openml\_637]{ 
\includegraphics[height=3.5cm, width=4cm]{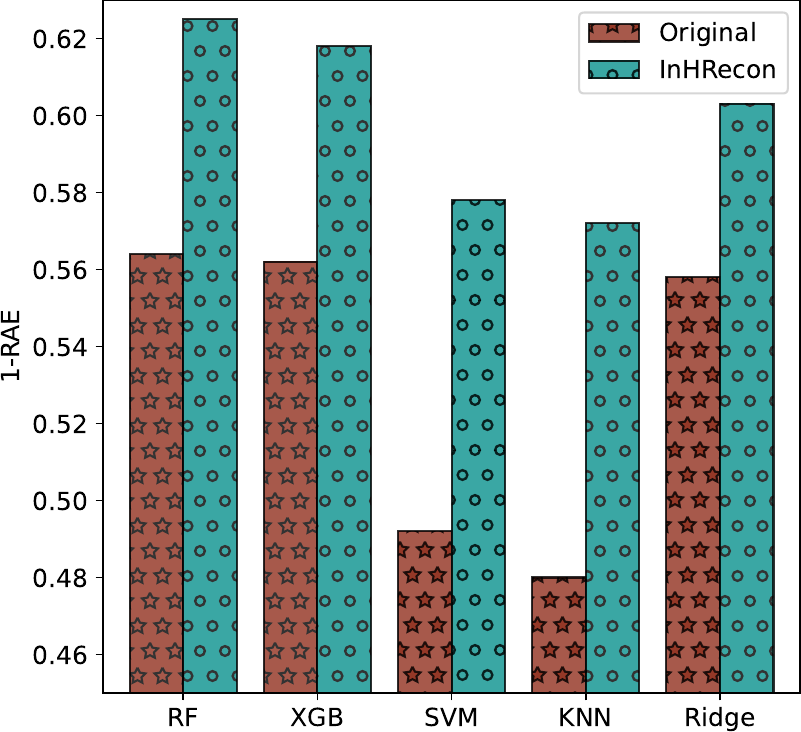}
}
\vspace{-0.35cm}
\caption{Comparison on different ML models in terms of F1 or 1-RAE.}
\label{differ_ml}
\vspace{-0.7cm}
\end{figure*}

\vspace{-0.4cm}
\subsection{Overall Performance} 
This experiment aims to answer: \textit{Can our proposed method construct optimal feature space to improve a downstream task?} \textbf{Table \ref{performance}} shows that, compared to six baselines, our model achieves state-of-the-art performance on 19 out of 24 datasets
overall. The underlying driver is that our personalized feature crossing strategy incorporates the strength of feature-feature interactions to generate new features. The superior performance of InHRecon over expansion-reduction-based (ERG, AFT) methods demonstrates the effectiveness of hierarchical sharing of states among agents, enabling optimal selection policies. Our self-learning end-to-end framework allows for easy application to diverse datasets, making it a practical and automated solution compared to state-of-the-art baselines (NFS, TTG) in real-world scenarios. Our model demonstrates a notable trend in its performance, showcasing superior results on real-world datasets such as \textit{PimaIndian} or \textit{German Credit}, compared to synthetic datasets like \textit{OpenML 620}. Synthetic datasets typically consist of features with randomly generated values, which might be better suited for AutoFE frameworks that rely on simple mathematical operations.

\vspace{-0.4cm}
\subsection{Ablation Study} This experiment aims to answer: \textit{How does each component in our model impact its performance?} We developed four variants of InHRecon (Section~\ref{baseline}). \textbf{Figure \ref{ab_study}} shows the comparison results
on two classification datasests (\textit{PimaIndian} and \textit{Credit\_Default}) and
two regression datasets (\textit{Housing Boston} and \textit{Openml\_637}). Unsurprisingly, InHRecon$^{-rnd}$ is consistently outperformed across all experiments. InHRecon surpasses InHRecon$^{-h}$, indicating that replicating expert cognition process facilitates the generation of meaningful and optimal features. Performances of InHRecon$^{-u}$ and InHRecon$^{-b}$ demonstrate that our personalized feature crossing strategy leads to improved feature space.

\vspace{-0.4cm}
\subsection{Study of Impact of H-statistics} This experiment aims to answer: \textit{Is H-statistics more effective than classical approaches in introducing statistical awareness in feature generation?} We replaced H-statistics with cosine similarity, Pearson's correlation, and mutual information measurements- aiming for the selected features to have lower redundancy and greater relevance to the prediction target. For instance, in experimental setup with MI, the utility metric is designed as follows:
\begin{equation}
    U(\mathcal{F}|y) = -\frac{1}{|\mathcal{F}|^2}\sum_{f_i, f_j \in \mathcal{F}} MI(f_i, f_j) + 
    \frac{1}{|\mathcal{F}|}\sum_{f\in \mathcal{F}}MI(f,y),
\end{equation}
where $\mathcal{F}$ represents the set of selected features $f_i$ and $f_j$, $|\mathcal{F}|$ the size of $\mathcal{F}$ and $y$ the target label. We report the comparison results on four different datasets. As seen from \textbf{Figure \ref{interact}}, H-statistics shows superiority across all datasets. The underlying driver is that, instead of merely calculating the degree to which two feature columns are related or how similar or dissimilar they are, H-statistics directly measures the share of variance that is explained by the interaction between two(or more) features. This allows us to prioritize the crossing of features that have more significant impact on prediction variation, resulting in faster convergence.
\vspace{-0.4cm}
\subsection{Robustness check of InHRecon under different ML models} 
This experiment aims to answer: \textit{Is InHRecon robust when different ML models are used in downstream task performance evaluation?} We examined the robustness of InHRecon by changing the ML model of a downstream task to Random Forest (RF), Xgboost (XGB), SVM, KNN, and Ridge Regression, respectively. The comparison results, depicted in \textbf{Figure \ref{differ_ml}}, demonstrate that InHRecon consistently enhances model performances across the tested datasets. This indicates that InHRecon exhibits strong generalization capabilities across various benchmark applications.


\vspace{-0.4cm}
\subsection{Parameter Sensitivity Analysis of InHRecon}
\begin{figure}[!t]
\subcapraggedrighttrue
\subcaphangtrue
\centering
\subfigure[Effect of higher order features]{
\includegraphics[height =3.5cm, width=4cm]{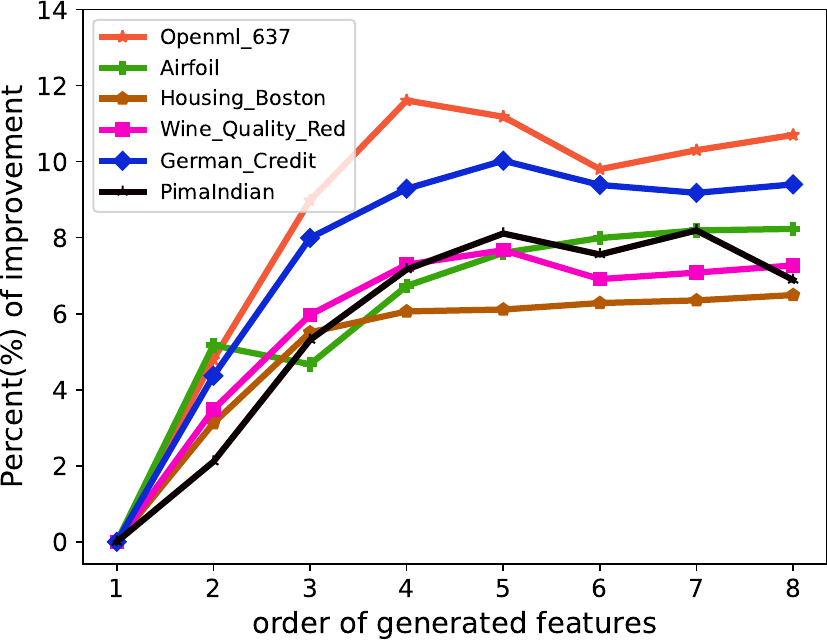}
}
\hspace{-3mm}
\subfigure[Effect of enlargement factor of feature space]{ 
\includegraphics[height =3.5cm, width=4cm]{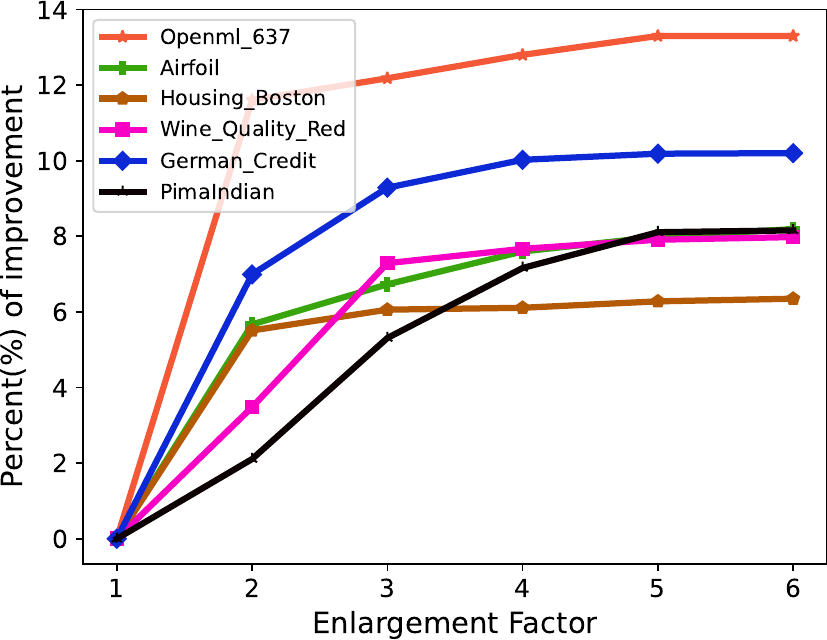}
}
\vspace{-0.3cm}
\caption{Parameter Sensitivity Analysis of InHRecon}
\label{param}
\vspace{-0.7cm}
\end{figure}

This experiment aims to answer: \textit{How InHRecon behaves under different parameter settings,} specifically the order of generated features and the enlargement factor of feature space. While the baseline methods simply overlook this, we argue that higher-order features are less interpretable to human observers. Hence, we report our performance results with the highest feature order set at 4. The results, as depicted in \textbf{Figure~\ref{param}}, indicate that InHRecon stabilizes around the 3rd to 5th order of features. We demonstrate that InHRecon achieves fast convergence when enlarging the feature space, typically only by a factor of 2x to 3x compared to the original feature space. 
These findings highlight the efficacy of our statistically aware feature crossing strategy, which intelligently explores the feature space and efficiently generates informative features.

\vspace{-0.4cm}
\subsection{Case Study: Rationality and Interpretability Analysis} This study aims to answer: \textit{Can InHRecon generate a rational and interpretable feature space?} In our \textit{German Credit} dataset case study, we use random forest classifier to identify the top 10 essential features for predicting credit risk. This dataset poses challenges in understanding due to its vague categorization, where feature columns are labeled with numbers 1 through 24. \textbf{Figure ~\ref{case_study}} presents the model performances in the central parts of each sub-figure, with corresponding features in the associated pie charts. Pie chart size corresponds to feature importance. We show that the \textbf{InHRecon}-reconstructed feature space enhances model performance by 5.75\%, with 40\% of the top 10 features being generated. This suggests that InHRecon generates informative features, leading to the refinement of feature space. Furthermore, our analysis highlights the impact of categorizing the features and expanding operation set to handle both numerical and categorical features.


%% file: related.tex
\section{Related Work} 

\noindent\textbf{Hierarchical Reinforcement Learning(HRL).} HRL enables simultaneous learning at multiple resolutions to accelerate the learning process~\cite{dayan1992feudal}. Hierarchical structures incorporate sub-goal information~\cite{parr1997reinforcement}, enhancing existing options~\cite{sutton1999between}. Value functions can be decomposed into individual sub-goals, facilitating improved exploration and learning efficiency~\cite{dietterich1998maxq} and decision-making abstractions~\cite{konidaris2019necessity}. In~\cite{tessler2017deep}, a deep HRL framework in lifelong learning is proposed, while ~\cite{nachum2018data} introduces HRL with learned goals for conveying instructions between policy levels. However, none of these methods are not suitable to be directly applied to learn strategies for AutoFE.

\begin{figure}[!t]
\subcapraggedrighttrue
\subcaphangtrue
\centering
\subfigure[Original Feature Space]{
\includegraphics[height =3.5cm, width=4cm]{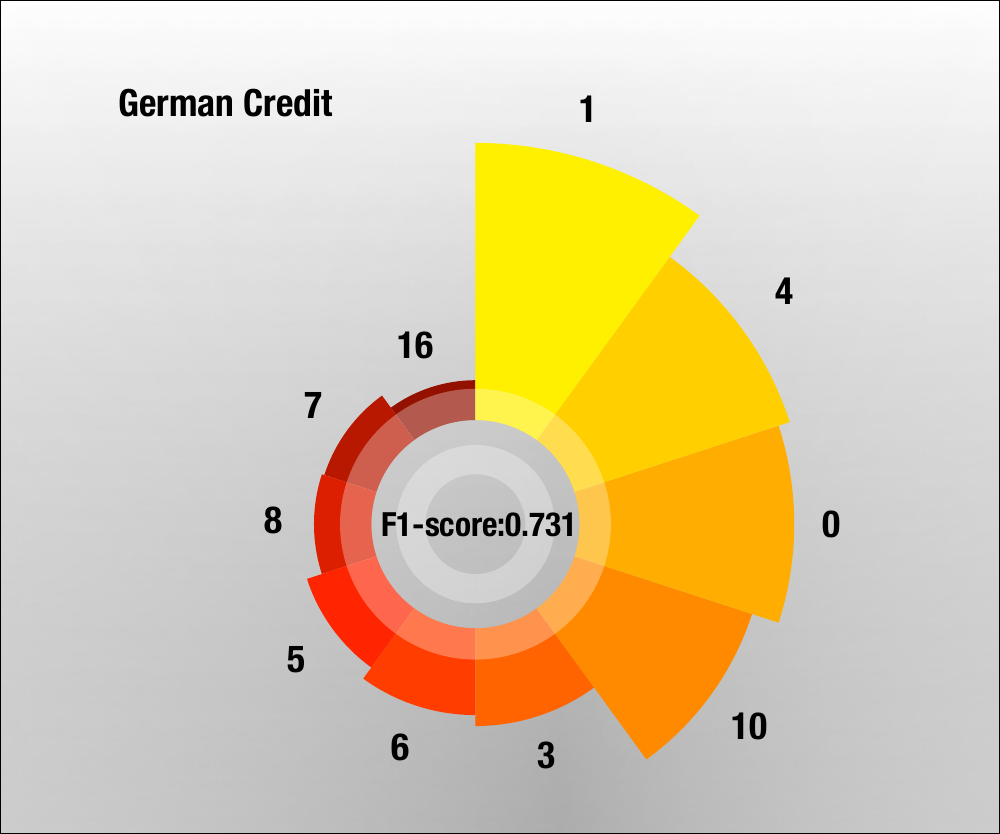}
}
\hspace{-3mm}
\subfigure[InHRecon-reconstructed Feature Space]{ 
\includegraphics[height =3.5cm, width=4cm]{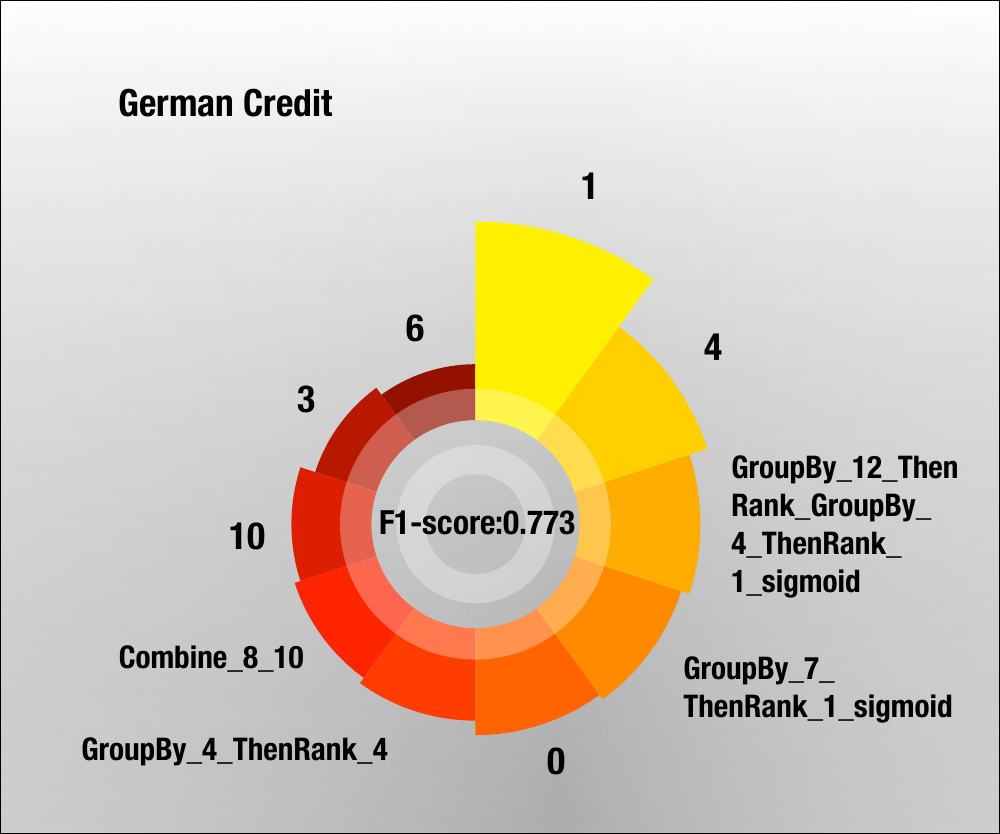}
}
\vspace{-0.2cm}
\caption{Top10 features for prediction in the original and InHRecon-reconstructed feature space}
\label{case_study}
\vspace{-0.6cm}
\end{figure}

\noindent\textbf{Automated Feature Engineering(AutoFE).} AutoFE aims to improve ML model performance by enhancing feature space through feature generation and selection techniques. \textbf{{\textit{Feature selection}}} eliminates redundancy and preserves important features, utilizing filter(e.g., correlation-based selection~\cite{yu2003feature}), wrapper(e.g., RL~\cite{liu2019automating, zhao2020simplifying, fan2020autofs, liu2021automated, liu2021efficient, liu2023interactive, xiao2023beyond} etc.), and embedded methods (e.g., decision tree~\cite{fan2021interactive} etc.). \textbf{{\textit{Feature generation}}} on the other hand, includes latent representation learning\cite{guo2017deepfm} and feature transformation approaches~\cite{chen2019neural,wang2021reinforced,khurana2018feature, 10.1145/3534678.3539278, xiao2023traceable, xiao2023traceable2, wang2023reinforcementenhanced}. These methods lack human-like and statistical awareness, leading to mindless exploration in larger feature spaces and causing inefficiency. Our personalized feature crossing captures highly relevant and interacted features and hierarchical agents learn effective interaction policies- all of which accelerate feature generation.

%% file: conclusion.tex
\section{Concluding Remarks}
We introduce \textit{InHRecon}, an interaction-aware hierarchical reinforced feature generation framework for optimal, interpretable and meaningful representation space reconstruction. We extend the operation space for RL agents, enabling them to emulate human expertise across various feature types. We decompose the feature generation process into sub-problems of operation selection and feature selection, addressed by hierarchical RL agents. We incorporate H-statistics as a feature interaction strength measurement to promote systematic exploration of the feature space and faster convergence. InHRecon also offers traceable feature generation, irmporving explainability.  Our framework achieves (surpassing or on-par) state-of-the-art performances on the standardized benchmarks adopted by prior work. 